\pdfoutput=1
\documentclass[10pt,final,conference,letterpaper]{ieeeconf}
\IEEEoverridecommandlockouts 
\usepackage{amsmath}

\DeclareMathOperator*{\argmax}{arg\,max}
\usepackage{amsfonts}
\usepackage{pgf}
\usepackage{tikz}
\usepackage{bm}
\usepackage{enumerate}
\usepackage[T1]{fontenc}
\usepackage[noadjust]{cite}
\usepackage{subcaption}
\usepackage{algorithmic}
\usepackage{array,multirow,graphicx}
\usepackage[ruled, vlined, plain,linesnumbered]{algorithm2e}
\usepackage{fancyhdr}

\title{\LARGE \bf Driving Style Encoder: Situational Reward Adaptation for General-Purpose Planning in Automated Driving}
\newcommand{\STAB}[1]{\begin{tabular}{@{}c@{}}#1\end{tabular}}
\author{Sascha Rosbach$^{1,2}$, Vinit James$^1$, Simon Gro{\ss}johann$^1$, Silviu Homoceanu$^1$, Xing Li$^1$ and Stefan Roth$^2$
\thanks{$^{1}$The authors are with the Volkswagen AG, 38440 Wolfsburg, Germany
        {\tt\small \{sascha.rosbach, vinit.james, simon.grossjohann, silviu.homoceanu, xing.li\}@volkswagen.de}}%
\thanks{$^{2}$The authors are with the Visual Inference Lab,
        Department of Computer Science, Technische Universit\"at Darmstadt,
        64289 Darmstadt, Germany
        {\tt\small stefan.roth@visinf.tu-darmstadt.de}}%
}
\begin{document}
\maketitle
\thispagestyle{fancy} 

\begin{abstract}
  General-purpose planning algorithms for automated driving combine mission, behavior, and local motion planning.
  Such planning algorithms map features of the environment and driving kinematics into complex reward functions.
  To achieve this, planning experts often rely on linear reward functions.
  The specification and tuning of these reward functions is a tedious process and requires significant experience.
  Moreover, a manually designed linear reward function does not generalize across different driving situations.
  In this work, we propose a deep learning approach based on inverse reinforcement learning that generates situation-dependent reward functions.
  Our neural network provides a mapping between features and actions of sampled driving policies of a model-predictive control-based planner and predicts reward functions for upcoming planning cycles.
  In our evaluation, we compare the driving style of reward functions predicted by our deep network against clustered and linear reward functions.
  Our proposed deep learning approach outperforms clustered linear reward functions and is at par with linear reward functions with a-priori knowledge about the situation.
\end{abstract}

\section{Introduction}
Automated driving in urban environments requires intelligent decision making that scales over a variety of traffic situations.
A scalable approach needs to be able to address both structured and un-structured traffic as experienced in urban environments.
In model-based planning, a semantic description of the environment is encoded in the form of static and kinematic features.
The planning system has to translate this semantic description of the environment into safe and human-acceptable actions.
The underlying planning algorithm often relies on manually-tuned linear reward functions that encode the relevance of the predefined features.
Tuning such reward functions is a tedious task and is usually performed by motion planning experts.
As a result, the reward function is often considered to be a static external signal that does not depend on the driving situation~\cite{levine2018reinforcement}.
Manual tuning of the reward function becomes infeasible if a planning system is applied at scale and has to adopt a variety of driving styles.
In this paper, we propose a deep learning approach in which a neural network dynamically predicts reward functions based on constantly changing static and kinematic features of the environment.

\begin{figure}  
  \vspace{1.8mm}
  \centering
  \includegraphics[width=0.48\textwidth]{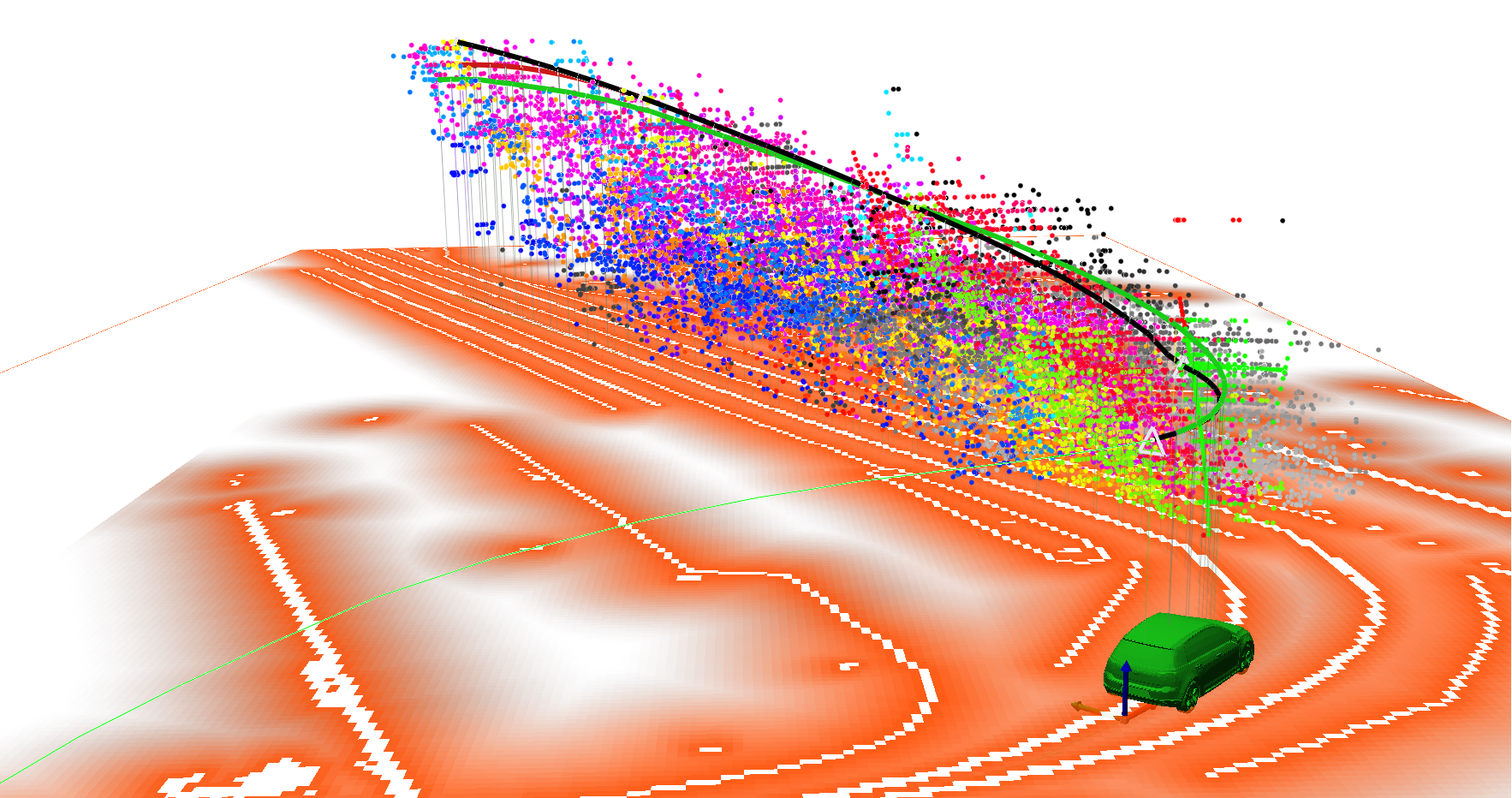}
  \caption{
    This figure illustrates our planner for automated driving that samples policies for our path integral (PI) IRL formulation.
    The visualized state space is color-coded based on the state-action values.
    The z-axis corresponds to the velocity, whereas the ground plane depicts the spatial feature maps such as distance transformed lane centers and road boundaries.
    Three color-coded policies are visualized, namely the optimal policy (black), the odometry of a human demonstration (green), and the projection of the demonstration into the state space (red).}
  \label{fig:title}
\end{figure}

Human driving demonstrations enable the application of inverse reinforcement learning (IRL) for finding the underlying reward functions~\cite{ng2000icml,kuderer2015icra}.
In this work, we utilize this methodology to learn situation-dependent reward functions for our planner~\cite{heinrich2015iros,rosbach2019,heinrich2018phd}.
Our planner samples a large set of actions generating distributions of path integral (PI) features and actions.
Unlike related work in deep inverse reinforcement learning (DIRL) that focuses on spatial reward functions \cite{wulfmeierMaximumEntropyDeep2015,wulfmeier2017ijrr}, our approach generates driving styles by incorporating vehicle kinematics.
This is done by integrating our deep learning approach into a model-predictive control (MPC) planning algorithm.
Sampled driving policies of the MPC are used as inputs to our neural network.
The network learns a representation of the driving situation by matching distributions of features and actions to reward functions based on the maximum entropy principle.

The main contributions of this paper are threefold:
First, we formulate a deep IRL methodology that predicts situation-dependent reward functions based on PI features and actions.
Second, we propose a neural network architecture that utilizes one-dimensional convolutions over PI features and actions of sampled driving policies to learn a representation of the statics and kinematics of the situation.
Third, we show the feasibility of our proposed approach in real-world traffic environments and compare our method to linear and latent maximum entropy IRL.

\section{Related Work}

As of now, the decision-making system in automated vehicles is often decomposed into a hierarchical structure to reduce the complexity of the model-based system. 
This hierarchical decomposition, however, yields uncertainty due to insufficient knowledge of the subsequent level.
In particular, these systems are prone to failure in unforeseen driving situations where simplified behavior models fail to match predefined templates~\cite{heinrich2018phd}.
General-purpose planning algorithms for automated driving aim to reduce planning-task decompositions to enable a scalable architecture that does not rely on behavior implementations.
Starting with the work of McNaughton~\cite{mcnaughton2011phd}, attention has been drawn to parallel real-time planning that combines behavior planning and local motion planning.
This approach enables sampling of a large set of actions that respect kinematic constraints.
Thereby a sequence of sampled actions can represent complex maneuvers.
This kind of general-purpose planner uses a linear reward function to map a complex feature space into rewards.
Manual tuning of such a linear reward function is a tedious task performed by motion-planning experts.
In our previous work, we automated the tuning process using IRL~\cite{rosbach2019}.
We found that inferred reward functions for situations defined a-priori surpass the performance of an expert-tuned reward function.
However, the tuned linear reward functions do not generalize well over different situations as the objectives change continuously, e.g., the importance of keeping to the lane center in straight road segments while allowing deviations in curvy parts.
In this work, we utilize the general-purpose planning approach and focus on situation-dependent reward function prediction.

A naive solution to the aforementioned problem is to use IRL with a-priori knowledge of situations and recover a set of situation-dependent reward functions.
However, the complexity is limited to the number of identifiable scenarios that have to be mapped during inference.
Prior work in IRL employs the latent maximum-entropy principle to condition reward functions on the situation~\cite{wangLatentMaximumEntropy2002}.
Expectation Maximization (EM) is often used in combination with IRL to learn a prior over the situation~\cite{babesApprenticeshipLearningMultiple2011}.
Training can be performed on a mixture of different driving segments without a-priori knowledge about the situation.
Hence, multiple reward functions are derived and a mixture is inferred during planning using Bayes' rule.
Dirichlet process mixture models have been used to address some drawbacks of EM, specifically the reliance on a specified number of clusters~\cite{shimosakaPredictingDrivingBehavior2015,choiNonparametricBayesianInverse2012}.
Recent work in IRL utilizes deep neural networks to extend the model capacity of mixture models.
Wulfmeier et al.~\cite{wulfmeierMaximumEntropyDeep2015} proposed to learn spatial traversability by deep IRL using deep convolutional neural networks, which generates a direct mapping from raw sensory data to reward maps.
This allows for encoding a situation into the reward function for a planning algorithm that operates on a grid-based representation.
However, their IRL formulation relies on state-visitation frequency updates that do not scale to high-dimensional state-action spaces.
 
In contrast to our model-based IRL formulation, prior work has also addressed high-dimensional state-spaces with model-free approaches.
Finn et al.~\cite{finnGuidedCostLearning2016} proposed a sample-based deep IRL approach that learns the cost in the inner loop of a model-free policy optimization.
This approach is suitable for problems with unknown system dynamics.
Recent work also includes the idea of combining generative adversarial neural networks and imitation learning to clone the behavior of the demonstrated policy without finding the underlying reward functions~\cite{hoGenerativeAdversarialImitation2016}.

In our work, we rely on a defined environment and vehicle transition model to explicitly incorporate traffic rules and provide safety~\cite{hruschka2019risk}.
We propose to couple the reliability of a model-based planning system with the generalization ability of deep inverse reinforcement learning.
Our search-based planning algorithm operates in large and continuous state space and produces policies with multiple behaviors.
This enables us to approximate the partition function required for the IRL gradient formulation with high accuracy~\cite{rosbach2019}.
Our IRL formulation uses a neural network to reason about static and kinematic features of the environment, thereby predicting situation-dependent driving styles.

\section{Preliminaries}
The interaction of the agent with the environment is often formulated as a Markov Decision Process (MDP), consisting of a 5-tuple \{$\mathcal{S}, \mathcal{A},T, R, \gamma$\}, where $\mathcal{S}$ denotes the set of states and $\mathcal{A}$ describes the set of actions.
The reward function $R$ is discounted by $\gamma$ and assigns a reward for every action $a \in \mathcal{A}$ in state $s \in \mathcal{S}$.
Our planning algorithm relies on an environment model $M$ and an underlying vehicle transition function $T$.
The model consists of static features of the environment that are derived from perception and localization, as well as kinematic features derived from the vehicle transition function.
The planner generates a policy set $\Pi$ by sampling actions $a$ from distributions conditioned on vehicle dynamics for each state $s$.
The features for each continuous action $a$ are integrated in the policy generation process.
The reward function $R$ is given by a linear combination of $K$ static and kinematic feature values $f_i$ with weights $\theta_i$ such that $R(s,a)=\sum_{i \in K}{-\theta_i f_i(s,a)}$.
The value $V^{\pi}$ of a policy $\pi$ is the integral of discounted rewards during continuous transitions, $V^{\pi}=\int_t \gamma_t R(s_t,a_t)\,dt$.
The PI feature $f^{\pi}_i$ for a policy $\pi$ is defined by ${f_i^{\pi}=\int_t\gamma_t f_i(s_t,a_t)\,dt}$.
The demonstrations for our IRL formulation are based on odometry records $\zeta$ of human driving.
During the training in our simulation, we project the odometry record $\zeta$ into the state-action space and are thereby able to formulate a demonstration from the policy set such that $\pi^D \in \Pi$.

\subsection{Maximum entropy IRL}

The goal of IRL is to find the reward function weights $\bm{\theta}$ that enable the optimal policy $\pi^*$ to be at least as good as the demonstrated policy $\pi^D$~\cite{arora2018}.
Thereby, the planner indirectly imitates the behavior of a demonstration~\cite{ng2000icml}.
In PI IRL, a probabilistic model is formulated that yields a probability distribution over policies, $p(\pi|\bm{\theta})$~\cite{aghasadeghi2011iros,theodorou2010generalized}.
The model is optimized such that the expected PI feature values $\mathbb{E}_{p(\pi|\bm{\theta})}[\bm{f}^{\pi}]$ of the policy set $\Pi$ match the empirical feature values $\hat{\bm{f}}^{\Pi^D}$ of the demonstrations for each planning cycle of the MPC.
Ziebart et al.~\cite{ziebart2008aaai} propose to maximize the entropy of the distribution to solve the ambiguity introduced by imperfect demonstrations, giving rise to the policy distribution
\begin{align}
  \label{eq:policyprob}
  p(\pi|\bm{\theta})= \dfrac{1}{Z}\exp(-\bm{\theta}^\top\bm{f}^{\pi}).
\end{align}

The calculation of the partition function $Z=\sum_{\pi \in \Pi}{\exp(-\bm{\theta}^\top \bm{f}^{\pi})}$ is often intractable due to the exponential growth of the state-action space.
Our planning algorithm approximates the partition function similar to Markov chain Monte Carlo methods.
Maximizing the entropy of the distribution over policies subject to the feature constraints from demonstrated policies implies that the log-likelihood $L(\bm{\theta})$ of the observed policies under the maximum entropy distribution is maximized.
From this hypothesis, the log-likelihood gradient is obtained as
\begin{equation}
  \nabla L(\bm{\theta}) = \sum_{\pi \in \Pi} p(\pi|\bm{\theta})\bm{f}^\pi - \hat{\bm{f}}^{\Pi^D}.
  \label{eq:linearirlgrad}
\end{equation}
Due to the dependency of the reward function on the driving situation, the probabilistic model $p(\pi|\bm{\theta})$ that recovers a single reward function for the demonstrated trajectories does not scale.

\section{Path Integral Latent IRL}
Instead of a single linear reward function, we consider that there are $N$ different reward functions, each corresponding to situation-dependent behavior.
As extension of our previous work, we incorporate latent variables in our PI IRL formulation~\cite{rosbach2019}, which allows us to infer a mixture of these reward functions during planning.
In the following derivation, we consider a single demonstration $\pi^D$ for every planning cycle.
We adopt the formulation of Babes et al.~\cite{babesApprenticeshipLearningMultiple2011} that utilizes Expectation Maximization (EM) in IRL, where $\bm{\beta}_c^{\pi^D}$ is the probability that a demonstration $\pi^D$ belongs to a cluster $c$, and $\bm{\psi}(c)$ is the 
the estimated prior probability of a cluster $c$.
The EM algorithm iteratively alternates between an estimation step (E-step) and a maximization step (M-step).
Within the E-step, we compute the probability of a demonstration belonging to a cluster as
\begin{equation}
  \bm{\beta}^{t, \pi^D}_c = \frac{p(\pi^D| c, \bm{\theta}^t, \bm{\psi}^t)\bm{\psi}^t(c)}{\sum_{c \in C} p(\pi^D|c, \bm{\theta}^t,\bm{\psi}^t) \bm{\psi}^t(c)},
  \label{eq:beta}
\end{equation}
exploiting the constraint $\sum_{c \in C}\bm{\beta}^{\pi^D}_c = 1$.
In Eq.~\ref{eq:beta}, $p(\pi^D|c,\bm{\theta},\bm{\psi})$ reduces to $p(\pi^D|\bm{\theta}_c)$.

Within the M-step, we compute for each iteration $t$ the prior probability $\bm{\psi}(c)$ as
\begin{equation}
  \bm{\psi}^{t+1}(c)=\frac{1}{B} \sum_{\pi^D \in \Pi^B} \bm{\beta}^{t,\pi^D}_c, 
\end{equation}
where $B$ is the number of demonstrations in a batch $\Pi^B$.
Furthermore, we compute the reward functions $\bm{\theta}_c$ for each cluster as
\begin{equation}
\bm{\theta}^{t+1}_c = \argmax_{\bm{\theta}} \sum_{\pi^D \in \Pi^B} \bm{\beta}^{t,\pi^D}_c \Big[\ln p(\pi^D|\bm{\theta}_c)\Big].
\end{equation}
We obtain the log-likelihood gradient for a cluster $c$ as
\begin{equation}
  \nabla_{\bm{\theta}_c} L(\bm{\theta}_c) = \bm{\beta}^{\pi^D}_c \Big(\sum_{\pi \in \Pi}  p(\pi|\bm{\theta}_c)\bm{f}^\pi - \bm{f}^{\pi^D} \Big).
  \label{eq:emirlgrad}
\end{equation}
In contrast to the gradient formulation of IRL that finds a single reward function for the demonstrated policies, latent IRL finds a number of reward functions.
The probability that a demonstration belongs to cluster $\bm{\beta}^{\pi}_c$ influences the update of the gradient.
During training, $\bm{\beta}^{\pi}_c$ is inferred using Bayes' rule on the basis of the PI features of the demonstration. 
Due to the absence of features of the demonstration during online planning, we assume that the optimal solution on the basis of the last predicted reward function allows us to infer a situation-dependent reward function for the next planning cycle.
In order to scale over a large number of situations, we next formulate an alternative approach based on deep learning.

\section{Path Integral Deep IRL}

We propose a deep learning approach for PI maximum entropy IRL that approximates a complex mapping between the situation and reward function.
Our neural network predicts the reward function weights $\bm{\theta}_{k+1}$ based on the actions $\bm{a}^{\Pi}_k$ and PI features $\bm{f}^{\Pi}_k$ of the policy set  $\Pi_k$ at MPC cycle $k$, given by $\bm{\theta}_{k+1} \approx g(\bm{\Theta},\bm{f}_k,\bm{a}_k)$.

The IRL problem can be formulated in the context of Bayesian inference as MAP estimation, which entails maximizing the joint posterior distribution of observing expert demonstrations $\Pi^D$.
We calculate the maximum entropy probability based on the linear reward weights $\bm{\theta}$, which are inferred by the network with parameters $\bm{\Theta}$ as
\begin{equation}
  L(\bm{\theta}) = L(g(\bm{\Theta},\bm{f},\bm{a})) =\sum_{\pi^D \in \Pi^D}\ln{p(\pi^D|g(\bm{\Theta},\bm{f},\bm{a}))}.
  \label{MLE}
\end{equation}

Maximization leads to the optimal weights 
\begin{equation}
  \bm{\theta}^* =\argmax_{\bm{\theta}} \sum_{\pi^D \in \Pi^D}\ln{p(\pi^D|g(\bm{\Theta},\bm{f},\bm{a}))}.
  \label{MLE}
\end{equation}

The gradient calculation from PI maximum entropy IRL from Eq. \ref{eq:linearirlgrad} lends itself naturally towards training deep neural networks, where the gradient of the log-likelihood $L(\bm{\theta})$ can be calculated in terms of $\bm{\Theta}$ as
\begin{equation}
  \begin{split}
    \frac{\partial L}{\partial \bm{\Theta}} & =\frac{\partial L}{\partial \bm{\theta}} \cdot \frac{\partial \bm{\theta}}{\partial \bm{\Theta}}\\
    & = \Big[\sum_{\pi \in \Pi} p(\pi|\bm{\theta})\bm{f}^\pi - \bm{\hat{f}}^{\Pi^D}\Big] \cdot\frac{\partial}{\partial \Theta}{g(\bm{\Theta},\bm{f},\bm{a} )}.
  \end{split}
  \label{eq:deepirlgrad}
\end{equation}

The gradient is separated into the maximum entropy gradient in terms of $\bm{\theta}$ and the gradient of $\bm{\theta}$ w.r.t.~the network parameters $\bm{\Theta}$, which can be directly obtained via backpropagation in the deep neural network.

\subsection{Training methodology}

Algo.~1 describes the training algorithm used for PI maximum entropy deep IRL.
IRL training is often very time consuming since the MDP has to be solved with the current reward function in the inner loop of reward learning.
In our IRL formulation, we do not solve the MDP within the inner loop; instead we run our planning algorithm prior to training with a randomly initialized reward function $\bm{\theta}_0$ over training segments.
Thereby, we generate a buffer of planning cycles consisting of a set of policy sets $\Pi$ with corresponding PI features $f^{\Pi}$ and actions $a^{\Pi}$.
Due to the high-resolution sampling of actions, we ensure that there are policies that are geometrically close to human-recorded odometry and resemble human driving styles.
Therefore the task of IRL is to find the unknown reward function that increases the likelihood of these trajectories to be considered as optimal policies.
During the policy generation, we utilize a weighted Euclidean distance towards human driven trajectories $\zeta$ to generate demonstrations $\Pi^D$ for each policy set $\Pi$.
Similar to the reward weights $\bm{\theta}$, the network parameters $\bm{\Theta}$ are initialized with random values before starting the training.
The training loop is run for a predefined number of epochs, ensuring that the convergence metric, the expected value difference (EVD), reaches the desired threshold value.
For each epoch the training dataset is shuffled and divided into batches to perform mini-batch gradient decent.

\begin{figure}[t]  
	\vspace{1.5mm}
	\centering
	\includegraphics[width=\linewidth]{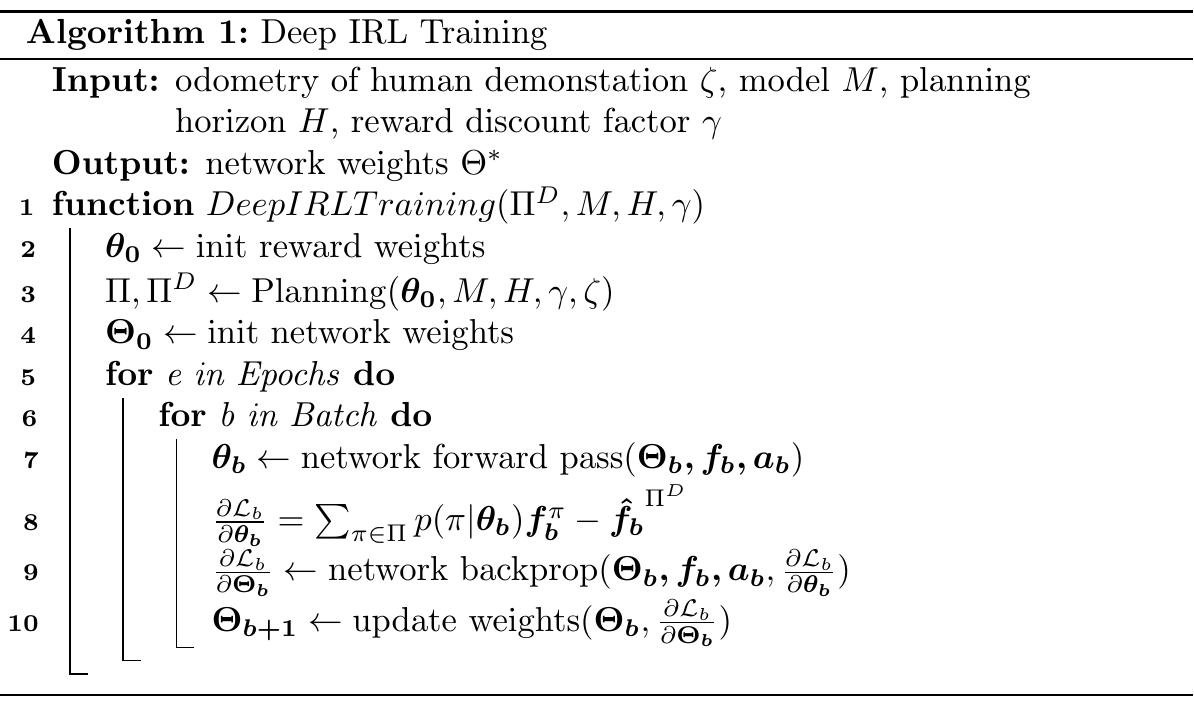}
\end{figure}

\subsection{Neural network architecture}

In this work a deep architecture is proposed, which maps the PI features $f^{\Pi}$ and actions $a^{\Pi}$ to linear reward weights $\bm{\theta}$.
The input to the network is a set of 15 PI features $f^{\Pi}$ and 8 actions $a^{\Pi}$ for every policy of a planning cycle.
The continuous actions are reduced to steering angle and accelerations at discrete control points.
The network architecture comprises of one-dimensional convolutions and average pooling to extract latent variables of policies describing the high-dimensional input space.
The architecture uses one-dimensional convolutional layers to learn the causal relationship between sampled actions and resultant features of every policy.
There is no inherent relationship among policies in the input space therefore no information is gained by performing higher dimensional convolutions.
The architecture consists of convolutional building blocks, each comprising of two convolutional layers followed by an average pooling layer which performs dimension reduction.
There are five such blocks in succession followed by fully-connected layers.
A series of eight fully-connected layers at the end learn inter-policy relationships from the resultant latent variables of the convolutional stack.
A vector of linear reward weights with the same dimension as the no.~of features, i.e.~15, is the output of the network.
All activation functions in the network are chosen as Rectified Linear Units, except for the output layer where no activation function is used.

\subsection{Inference during MPC planning cycles}
During real-time planning in an automated drive, the MPC re-plans in discrete time-steps $k$. 
After receiving the features and actions of the latest planning cycle, the neural network infers the reward weights.
To enable smooth transitions of the reward functions, we utilize a predefined history size $h$ to calculate the empirical mean of weights $\hat{\bm{\theta}}$.
The weights hence obtained are used to continuously re-parameterize the planning algorithm for the subsequent planning cycle.

\section{Experiments}
In our experiments, we use our proposed PI deep IRL (PIDIRL), as well as PI latent IRL (PILIRL) and PI linear IRL (PIIRL) in real urban driving situations.
We focus our experiments on situations that introduce conflicts in driving objectives so as to highlight the importance of situation-dependent reward functions.

\subsection{Data collection and simulation}
Our experiments are conducted on a prototype vehicle, which uses a mediated perception architecture to produce feature maps of the environment.
These feature maps are depicted in the ground-plane in Fig.~\ref{fig:title}.
The MPC planning algorithm uses the feature maps to get state features and computes state-values as shown in as color-coded point clouds in Fig.~\ref{fig:title}.
We gathered data on a set of streets in Hamburg, Germany located around Messe and Congress, see Fig.~\ref{fig:overview_situations}.
We concentrate on static infrastructure and objects while disabling intention prediction of dynamic objects to test our reward functions in different location-dependent situations on this track.
Three drives over this course are segmented into train and test tracks, each of which is subdivided into four situations.
Situation 1 resembles sharp curves, situation 2 stopping, starting and turns at traffic lights, situation 3 stopping and starting at a traffic lights, and situation 4 resembles lane following.
The training of our algorithm is performed in a playback simulation of the recorded data as depicted in Fig.~\ref{fig:title}.
After every planning cycle of the MPC, the position of the vehicle is reset to the odometry recording of the human demonstration

\subsection{Reward feature representation}
We concentrate on a reward set consisting of $K$ = 15 manually engineered features, which are listed in the table of Fig.~\ref{table:features}.
Infrastructural features are derived by a data fusion between objects and street network~\cite{homeier2011itsc}.
The kinematic characteristics of the policies are given by derivatives of the lateral and longitudinal actions as well as features related to behavior, e.g. lane change delay.
In addition selection features are designed so as to evaluate policies based on more nuanced attributes such as maneuvers space, progress and policy end direction toward the road. 

\section{Evaluation}
We compare the results of the situation-dependent reward functions from PI deep IRL (PIDIRL) against PI latent IRL (PILIRL), and PI linear IRL (PIIRL) with and without a-priori knowledge about the situation.
First, we compare the model training based on distance convergence toward the human demonstration.
Second, we compare the driving style of our policies under learned reward functions against human driving in different driving situations. 
Third, we analyze the reward weights inferred on the different test segments.

 \subsection{Training evaluation}

\begin{figure}[t]
  \vspace{2mm}
  \centering
  \includegraphics{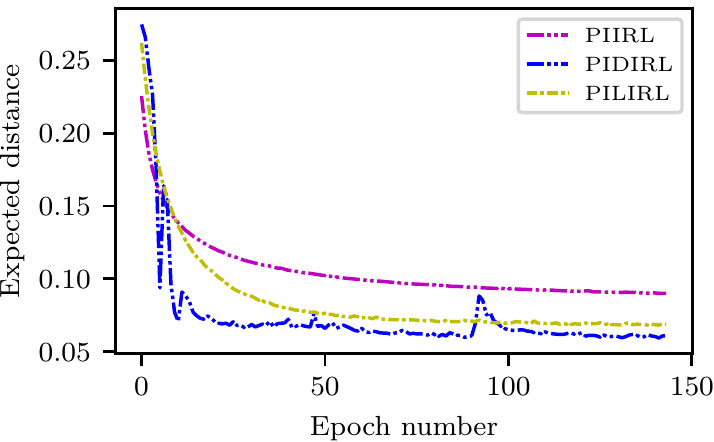}
  \caption{Expected distance over epochs of training.}
  \label{fig:edplot}
\end{figure}

We analyze the convergence of our model training based on a distance metric $d(\zeta,\pi)$ towards the human driven demonstration $\zeta$.
This distance metric is an integral of the Euclidean distance in longitudinal and lateral direction as well as the squared difference in the yaw angle over the policy as described in detail in~\cite{rosbach2019}.
Due to our goal of optimizing the likelihood of human driving behavior within the policy set $\Pi$, we measure our training convergence based on the expected distance (ED), given by $\mathbb{E}[d(\zeta,\Pi)]= \sum_{\pi \in \Pi} {p(\pi|\bm{\theta}) d(\zeta,\pi)}$.
We plot ED for each IRL approach over an equal amount of training epochs.
In Fig.~\ref{fig:edplot} the ED for PIDIRL, PILIRL and PIIRL are plotted for 143 epochs.
The PIDIRL shows a large ED reduction 78\% over the entire training dataset.
The starting ED is high due to the randomly initialized network, but a high reduction of 74\% is achieved in only 22 epochs.
PILIRL also has a high initial ED due to higher variance in weight initialization in order to find multiple different clusters, it finally shows a reduction of 74\%. 
Due to the limited model complexity, PIIRL shows lowest ED reduction of 60\%, here trained over the entire training dataset without a-apriori knowledge about the situations.
The slight variance in the PIDIRL plot, particular in epoch 90, can be explained due to high learning rate and small batch size.
From the statics above we can deduce that PIDIRL has the highest likelihood of selecting policies with humanlike driving behavior across different driving situations.

\subsection{Driving style evaluation}
\begin{figure*}
  \centering
  \vspace{2mm}
    \begin{subfigure}{0.45\textwidth}
            \begin{subfigure}{0.3\textwidth}
              \centering
              \begin{tabular}{c|c}
                Label & Description \\ \hline
                1 & Sharp Turn \\
                2 & Stop, Start, Turn\\
                3 & Stop, Start\\
                4 & Lane follow\\
                \end{tabular}
            \end{subfigure}
            \hfill
            \begin{subfigure}{0.7\textwidth}
              \centering
            \includegraphics[scale=0.25]{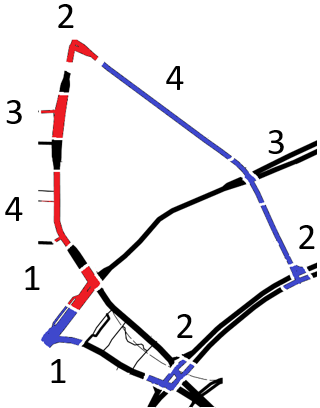}
            \end{subfigure}
            \caption{The map depicts the training (blue) and test (red) segments of the selected Route in Hamburg City Center.}
            \label{fig:overview_situations}
          \end{subfigure}
            ~
            \begin{subfigure}{0.45\textwidth}
              \centering
              \includegraphics[scale=0.9]{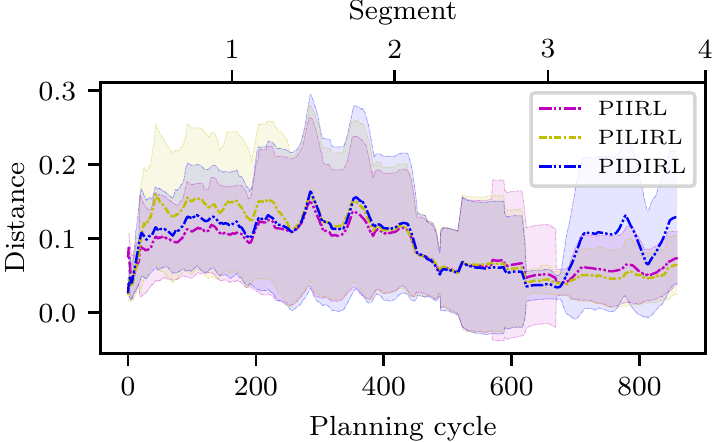}
              \caption{Distance of the opt. policies on different driving segments.}
              \label{fig:optcycles}
            \end{subfigure}
            \vspace{1mm}
            \begin{subfigure}{0.40\textwidth}
              \centering
              \includegraphics[width=0.8\textwidth]{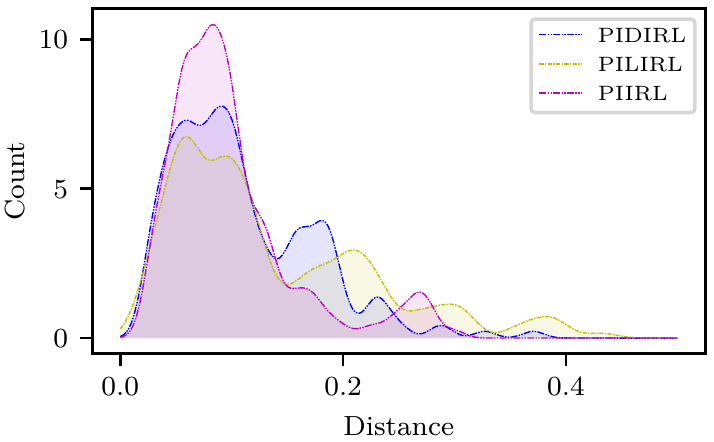}
              \caption{Distribution of the opt. policy distances in seg. 1.}
              \label{fig:hist1}
            \end{subfigure}
            \hspace{2mm}
            \begin{subfigure}{0.40\textwidth}
              \centering
              \includegraphics[width=0.8\textwidth]{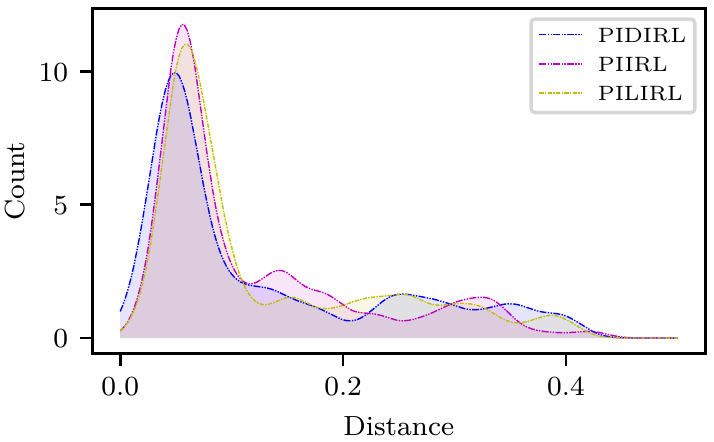}
              \caption{Distribution of the opt. policy distances in seg. 2.}
              \label{fig:hist2}
            \end{subfigure}
            \vspace{1mm}
            \begin{subfigure}{0.40\textwidth}
              \centering
              \includegraphics[width=0.8\textwidth]{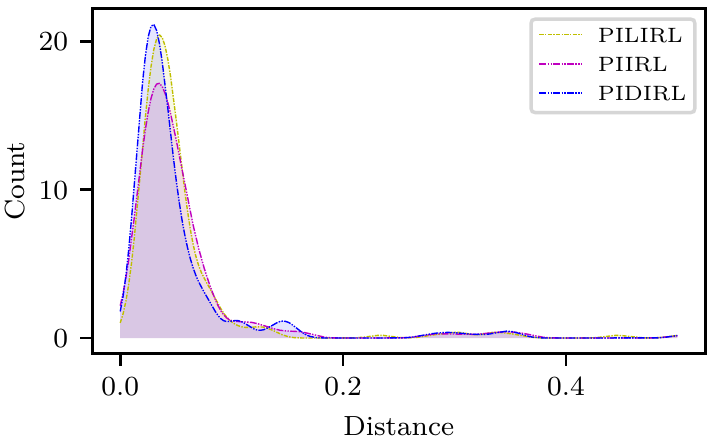}
              \caption{Distribution of the opt. policy distances in seg. 3.}
              \label{fig:hist3}
            \end{subfigure}
            \hspace{2mm}
            \begin{subfigure}{0.40\textwidth}
              \centering
              \includegraphics[width=0.8\textwidth]{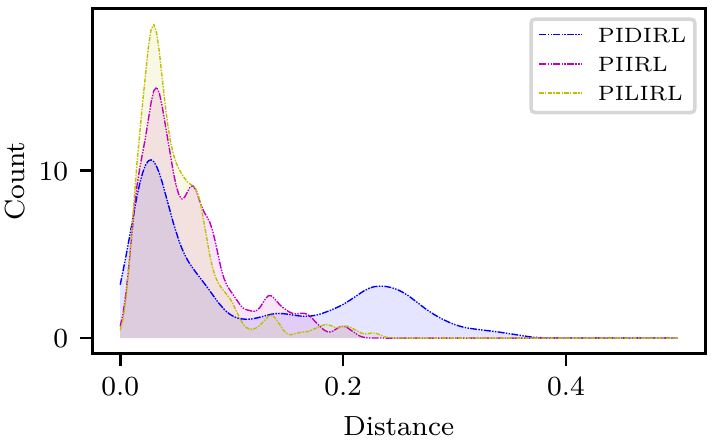}
              \caption{Distribution of the opt. policy distances in seg. 4.}
              \label{fig:hist4}
            \end{subfigure}
            \begin{subfigure}{0.4\textwidth}
                \begin{tabular}{c|c|l}
                  & \multicolumn{1}{c|}{Label} & \multicolumn{1}{l}{Feature} \\ \hline
                  \multirow{7}{*}{\STAB{\rotatebox[origin=c]{90}{Kinematics}}}
                  &         0 & \text{Long. Acc.} \\                                                                          
                  &         1 & \text{Long. Jerk} \\
                  &         2 & \text{Long. Velocity} \\
                  &         3 & \text{Lat. Acc.} \\
                  &         4 & \text{Lat. Jerk} \\ 
                  &         9 & \text{Lat. Overshooting} \\                                                              
                  &         10 & \text{Lane Change Delay} \\ \hline
                  \multirow{4}{*}{\STAB{\rotatebox[origin=c]{90}{Statics}}}
                  &         5 & \text{Centerline} \\
                  &         6 & \text{Direction} \\
                  &         7 & \text{Proximity} \\
                  &         8 & \text{Curbs} \\ \hline
                  \multirow{4}{*}{\STAB{\rotatebox[origin=c]{90}{Selection}}}
                  &         11 & \text{State Class} \\ 
                  &         12 & \text{Manuver Space} \\
                  &         13 & \text{End Direction} \\
                  &         14 & \text{Min. Progress} \\
                \end{tabular}
                \caption{The table lists the considered PI features.
                  The parameters are plotted in logarithmic scale.
                }
                \label{table:features}
                \begin{subfigure}{0.8\textwidth}
                  \vspace{1mm}
                  \includegraphics[width=\textwidth]{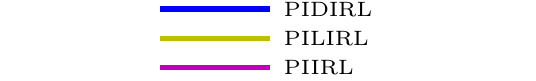}
                \end{subfigure}
            \end{subfigure}
            ~
            \begin{subfigure}{0.57\textwidth}
              \begin{subfigure}{\textwidth}
              \begin{subfigure}{0.43\textwidth}
                \centering
                \includegraphics[trim=25 25 25 25,clip,width=\textwidth]{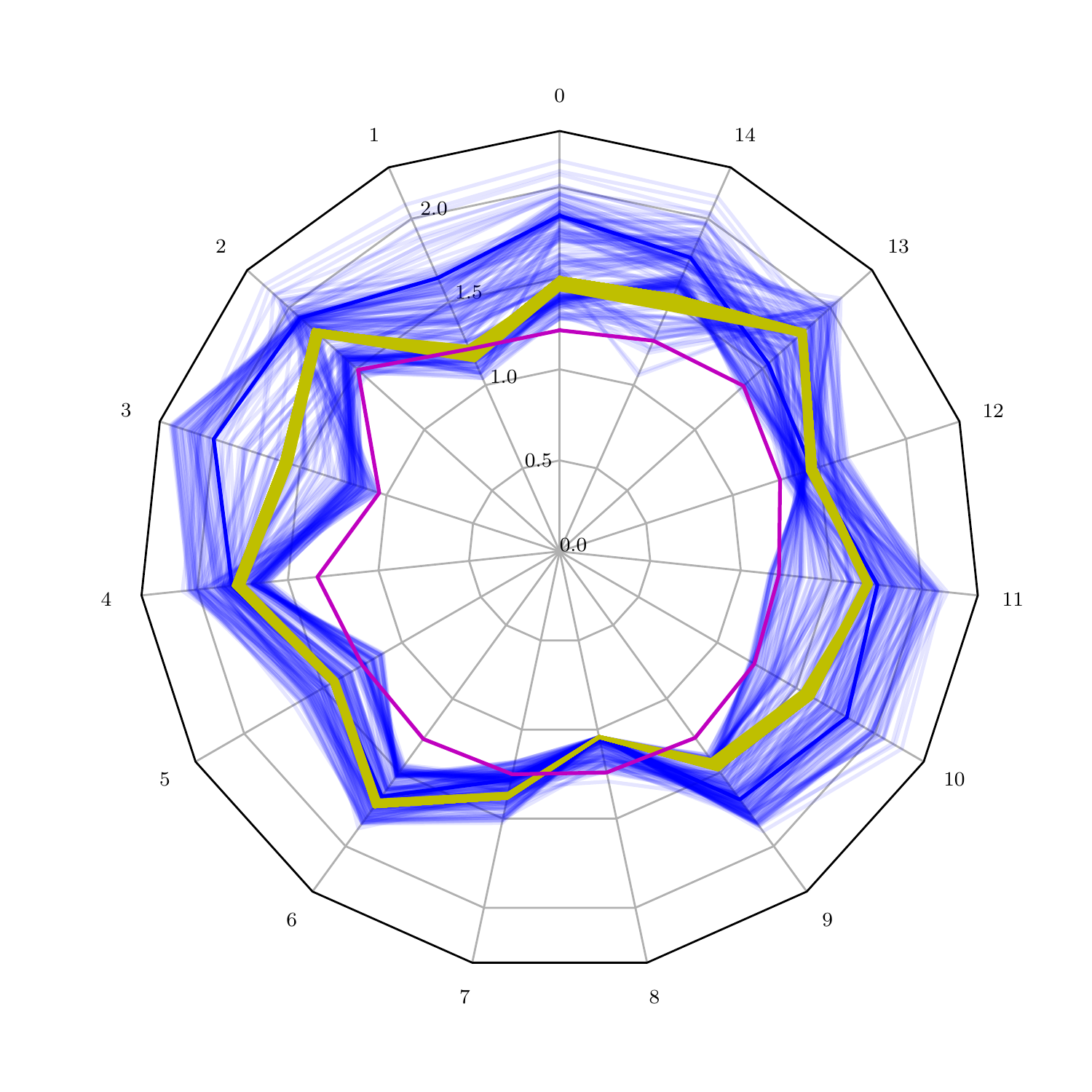}
                \label{fig:param1}
                \vspace{-6mm}
                \caption{Inf. reward weights on seg. 1.}
              \end{subfigure}
              ~
              \begin{subfigure}{0.43\textwidth}
                \centering
                \includegraphics[trim=25 25 25 25,clip,width=\textwidth]{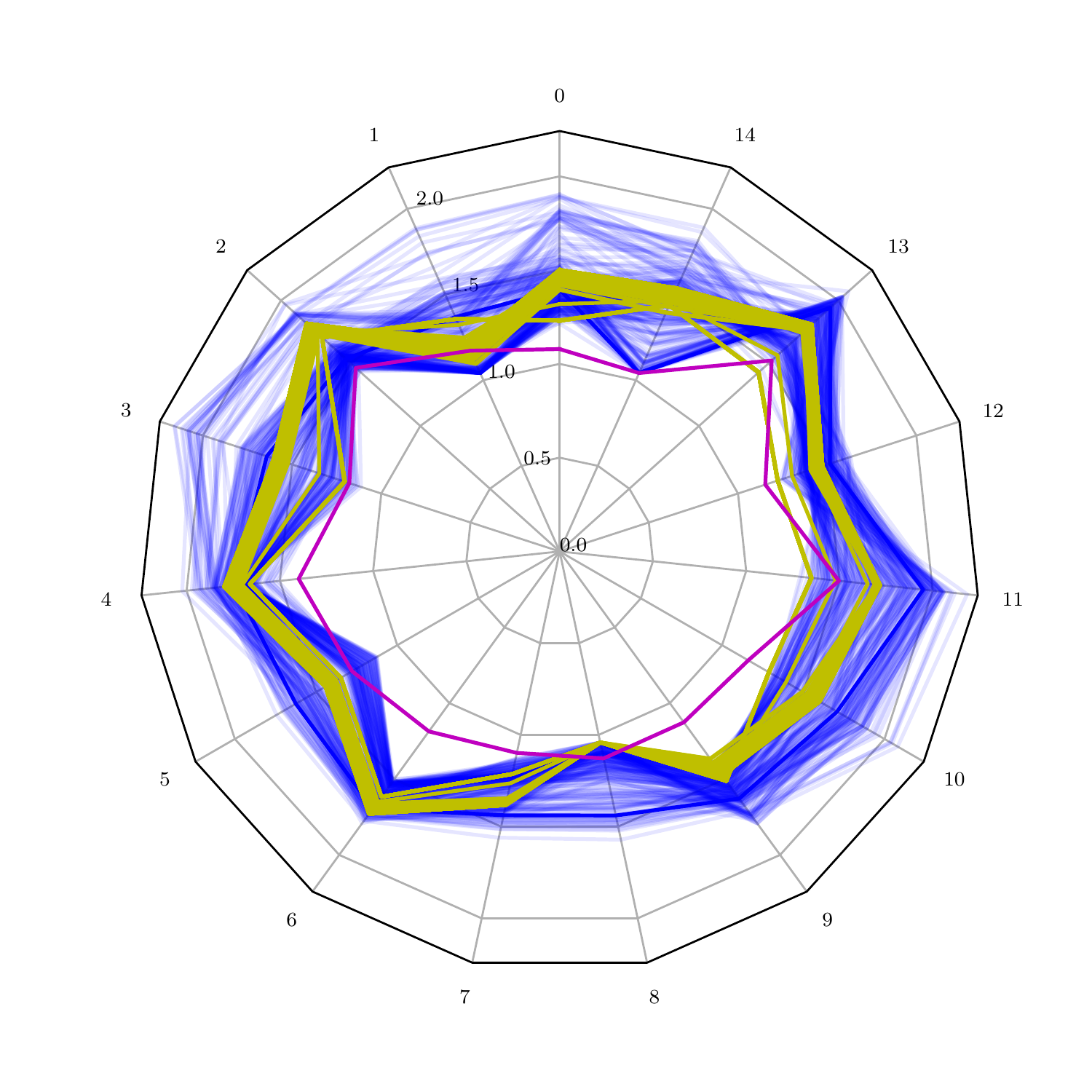}
                \label{fig:param2}
                \vspace{-6mm}
                \caption{Inf. reward weights on seg. 2.}
              \end{subfigure}
            \end{subfigure}
              \begin{subfigure}{\textwidth}
              \begin{subfigure}{0.43\textwidth}
                \centering
                \includegraphics[trim=25 25 25 25,clip,width=\textwidth]{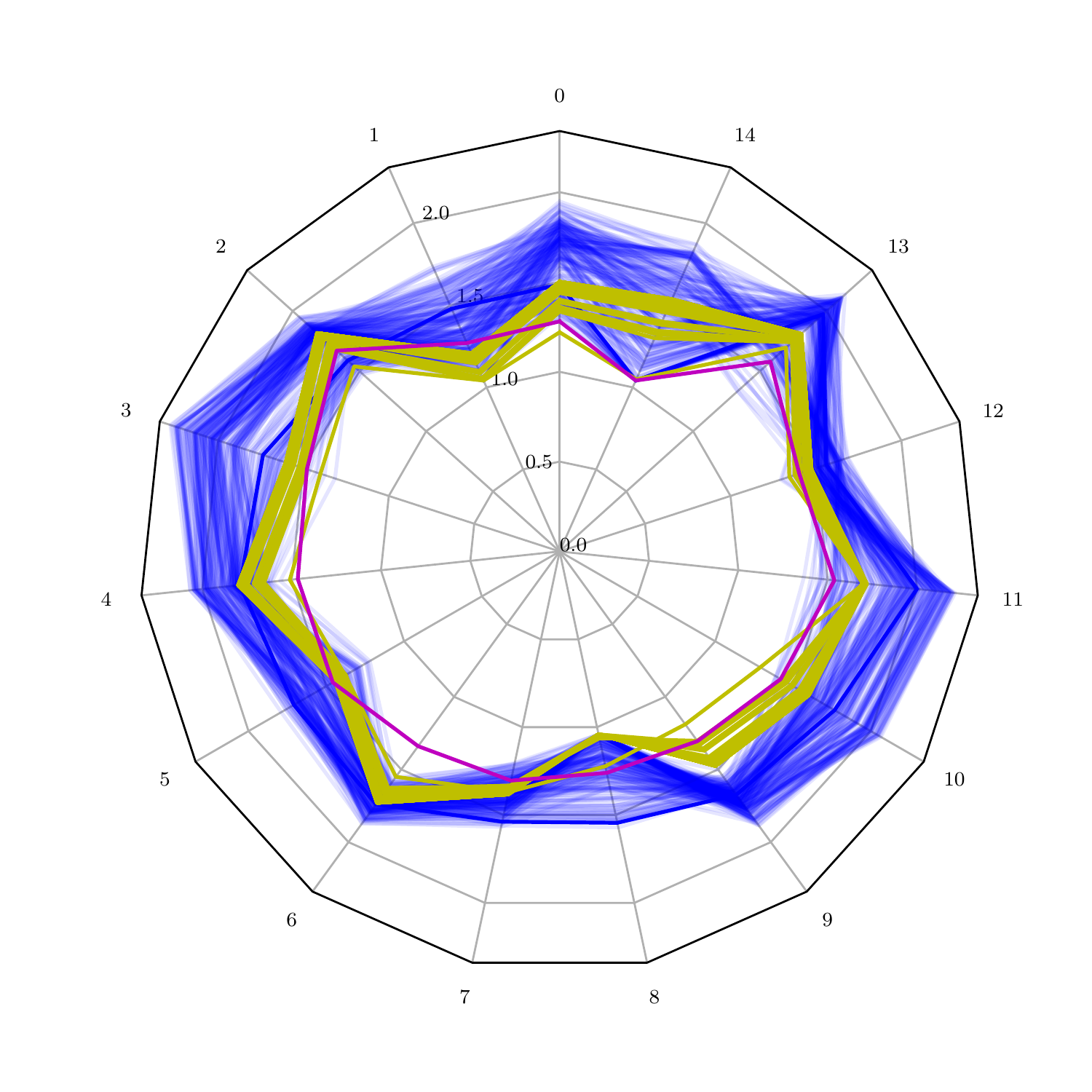}
                \label{fig:param3}
                \vspace{-6mm}
                \caption{Inf. reward weights on seg. 3.}
              \end{subfigure}
              ~
              \begin{subfigure}{0.43\textwidth}
                \centering
                \includegraphics[trim=25 25 25 25,clip,width=\textwidth]{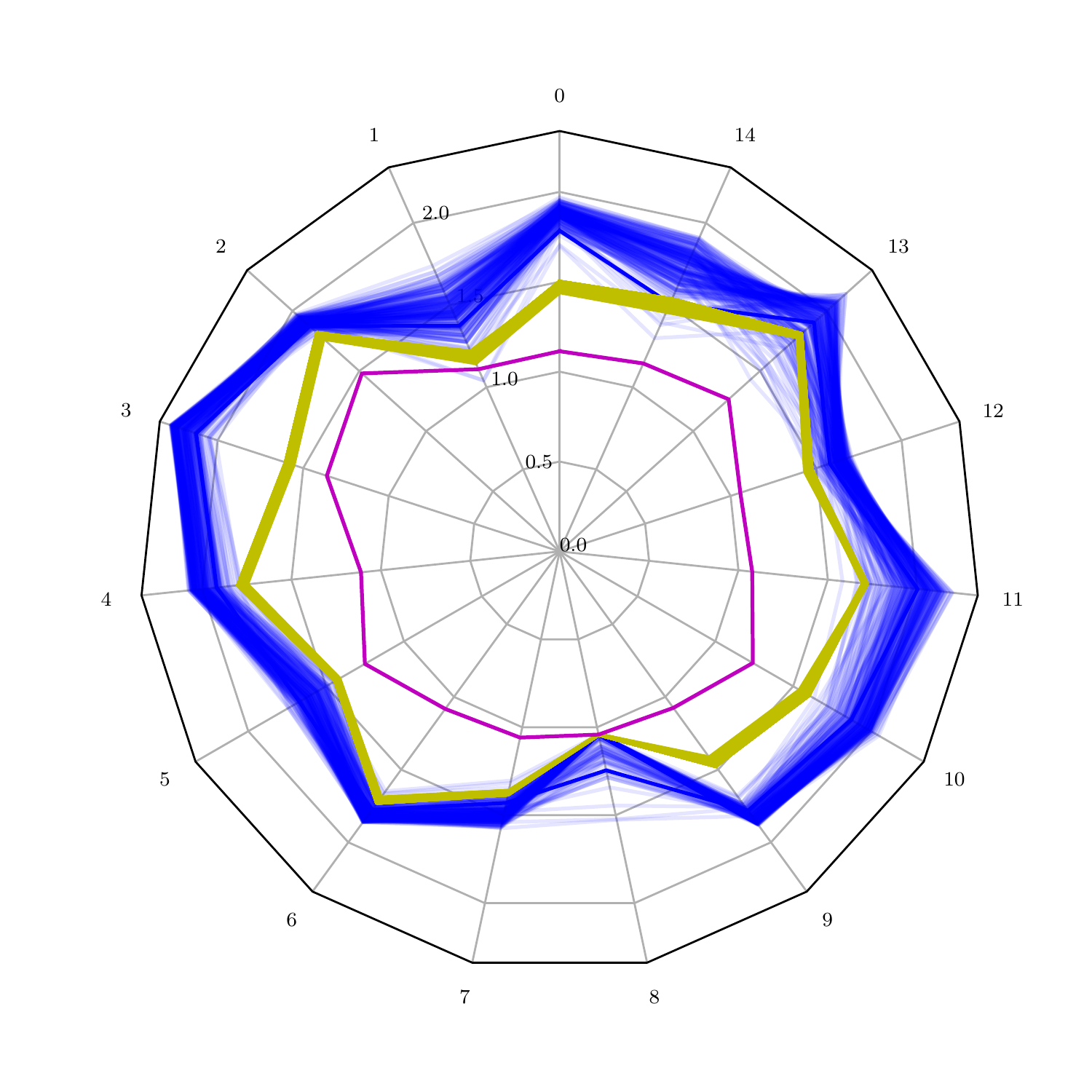}
                \label{fig:param4}
                \vspace{-6mm}
                \caption{Inf. reward weights on seg. 4.}
              \end{subfigure}
              \end{subfigure}
            \end{subfigure}
             \caption{
               \emph{(a)} Tests conducted on a selected Route in Hamburg City center.
               \emph{(b-f)} Graphs compares the performance of PIDIRL with PILIRL, and PIIRL over different driving segments.
               \emph{(g)} Table of reward function features.
               \emph{(h-k)} Graphs show the reward function predictions of our methods over driving segments.
             }
   \label{fig:test}
\end{figure*}

In this part of the evaluation, we compare the driving style of PIDIRL, PILIRL, and PIIRL models against manual human driving over different test situations.
We generate multiple PIIRL models with a-prior knowledge for each driving situations so as to compare the situation-dependent performance of the proposed PIDIRL.
We plot the distance of the optimal policy obtained from theses models to the odometry of the human demonstration for different driving situations.
Our test segments contain 187 planning cycles for segment 1, 232 for segment 2, 217 for segment 3, and 224 for segment 4.
In Fig.~3c-f, we fit a Gaussian distribution over the segment specific distance histogram with 200 bins of size 0.0025.
From our experiments we deduce that our proposed PIDIRL method produces policies with 21\% lower mean distance as compared to PILIRL in seg. 1 and in seg. 2.
In seg. 3 the mean distances from both of these methods are comparable.
Moreover, PIDIRL shows 13\% lower mean distance as compared to the segment specific PIIRL for seg. 3 and comparable results for the other segments.
In seg. 4 PIDIRL produces policies with higher mean distances as compared to PIIRL and PILIRL.
This trend can be described by the dominant prior of PILIRL and overfitting of PIIRL for straight segments.
This poses as a disadvantage for both of theses methods as compared to our proposed PIDIRL which has better generalization capabilities.
This generalization can be seen in the Fig.~\ref{fig:optcycles}, where PIDIRL has a similar mean distance over all segments.
The reward function weights obtained from these models over the test situations are shown in Fig.~3h-k, where the weights are inversely proportional to the preference of that feature value in the policy.
The reward weights are plotted in logarithmic scale.
In these plots, it can be seen that PIDIRL produces nuanced reward functions for every planning cycle in the segment.
The variance of the reward function produced by PIDIRL is proportional to the situation complexity.
In simple situations, as shown in Fig.~\ref{fig:optcycles} at seg. 4, having a nuanced reward functions prediction mechanism like PIDIRL producing high variance as shown in Fig.~3k could lead to lower performance as compared to simpler reward functions namely PILIRL and PIIRL.

 \section{Conclusion and Future Work}

We utilize path integral (PI) maximum entropy deep IRL to learn situation-dependent reward functions of a general-purpose planning algorithm.
We propose a method to couple the reliability of a model-based planning system with the generalization capability of deep inverse reinforcement learning.
Our experiments show that reward function predictions by our proposed neural network architecture are at par with multiple PI linear IRL reward function model that are trained with a-priori knowledge about the situation.
In addition to this, we show that our deep IRL methodology has better generalization capabilities as compared to PI latent IRL which uses behavioral clustering of demonstrations.
In future we plan to experiment with different network architectures in our proposed deep IRL paradigm so as to reduce the variance in reward functions prediction in simple driving situations.

\bibliographystyle{IEEEtran}
\bibliography{bib/conf_names_abrv,bib/library}
\end{document}